\documentclass{article}
\usepackage[final]{neurips_2026_vericode}   % camera-ready final mode
\usepackage[utf8]{inputenc}
\usepackage[T1]{fontenc}
\usepackage{hyperref,url,booktabs,amsmath,amssymb,graphicx,xcolor,microtype}
\providecommand{\doi}[1]{doi: \href{https://doi.org/#1}{\nolinkurl{#1}}}
\usepackage{listings}
\usepackage{float}
\graphicspath{{figures/}}

\newcommand{\nCompiling}{185}
\newcommand{\nCertified}{458}
\newcommand{\certTthree}{50/122}

\newcommand{\deltaGoedelAll}{+20.3}

\title{How Stable Are LLM-Generated Proofs Under Program Refactoring?\\ A Stress Test for Verifiable Code Generation}
\workshoptitle{AI for Verifiable Coding}

\author{%
  Aravind Mandiga \\
  University of Georgia \\
  Athens, USA \\
  \texttt{aravind.mandiga@uga.edu}
  \And
  Guoming Li \\
  University of Georgia \\
  Athens, USA \\
  \texttt{gmli@uga.edu}
}

\makeatletter\g@addto@macro\normalsize{\setlength\abovedisplayskip{4pt plus 1pt}\setlength\belowdisplayskip{4pt plus 1pt}\setlength\abovedisplayshortskip{2pt}\setlength\belowdisplayshortskip{2pt}}\makeatother
\begin{document}
\maketitle

\begin{abstract}
Benchmarks for verifiable code generation usually test a proof against one fixed implementation.
We ask whether the same proof remains valid after a program refactoring that preserves behavior.
We introduce \textbf{RefactorProof}, which derives \nCertified{} Lean-certified variants from the
\nCompiling{} VERINA tasks that compile in our pinned environment while holding the specification and
evaluated proof fixed byte-for-byte. VERINA supplies reference proofs for 46 tasks, all from its basic
split; the primary model-versus-reference comparisons further match on 21--25 short, non-recursive tasks.
Structural refactorings expose substantial brittleness across proof sources: task-normalized structural PSR
is 4.3\% for the supplied references and 13.3--26.8\% for the specialized provers. On the matched tasks,
specialized prover proofs nevertheless have higher overall unchanged-proof survival than the supplied
reference artifacts: +15.6 points for DeepSeek-Prover-V2-7B, +18.8 for Goedel-Prover-V2-32B, and
\deltaGoedelAll{} for Goedel-Prover-V2-8B; all three 95\% CIs exclude zero. The ordering persists after
excluding universally survived T1 variants, under family-balanced aggregation, and under alternative
deployable valid-proof selection rules. However, unchanged-proof failure is not the same as high repair
cost: predefined one-identifier edits repair 51/54 eligible reference structural failures and 58/85 eligible
prover failures. RefactorProof therefore measures unchanged-proof stability under program refactoring,
not the cost of maintaining a proof after it breaks.
\end{abstract}

\section{Introduction}
A machine-checked proof establishes that an implementation satisfies its specification. That guarantee is
strong, but it is usually evaluated on one fixed version of the program. Real programs are refactored, and
proofs can break even when those changes preserve behavior. Proof maintenance is a long-standing problem in
formal methods. A Lean user may expect some scripts that unfold definitions to break after helper extraction;
the empirical question is whether this brittleness differs systematically across proof sources and
transformation mechanisms. The new setting considered here is that language models now generate such proofs
at scale, while current verifiable-code benchmarks still score whether a proof verifies once.

We introduce RefactorProof to measure this gap. A VERINA task contains an implementation $C$, a specification
$S$, and a proof $P$. We mechanically construct a refactored implementation $C'$, ask Lean to certify
$\forall \bar x.\, C(\bar x)=C'(\bar x)$ in a separate file, and then re-check the unchanged proof $P$ against
$(C',S)$. The proof survival rate (PSR) is the fraction of certified refactorings of a task on which the same
proof still verifies. RefactorProof uses five deterministic transformation families grouped into three
operational classes: definitional ($D$), structural ($S$), and theorem-certified semantic ($E$).
These labels describe the source-level change being tested. They do not classify Lean's definitional
equality: T2 and T5 alter source structure even though Lean can reduce the transformed implementation
to the original behavior definitionally.

\paragraph{Contributions.}
(i) A semantics-preserving evaluation in which only the implementation changes, while the specification and
proof are byte-fixed and every retained transformation has a separate Lean certificate.
(ii) RefactorProof, a benchmark of \nCertified{} certified variants, together with matched proof-survival
comparisons and sensitivity analyses for proof selection, aggregation, common support, and certification
artifacts.
(iii) Artifact-level analysis of how proof strategies fail under different refactorings, plus a controlled
mechanical-repair experiment that separates unchanged-proof stability from simple repair effort.
We do not claim that proof brittleness itself is new. The supplied VERINA reference proofs are available
for only 46 basic-split tasks, and the primary model-versus-reference comparisons match on 21--25 short,
non-recursive tasks. We therefore interpret the source-level differences as comparisons of proof artifacts
on this subset, not as checkpoint-wide robustness rankings over VERINA. Code, benchmark artifacts, and
analysis scripts are available at
\url{https://github.com/aravind-mandiga/refactorproof}.

\section{Related work}
\paragraph{Benchmarks for verifiable code generation.}
Benchmarks that ask models for code with formal specifications and machine-checked proofs include
VERINA~\citep{ye2025verina}, which we build on, CLEVER~\citep{thakur2025clever} and
miniCodeProps~\citep{lohn2024minicodeprops} in Lean~4, FVAPPS~\citep{dougherty2025fvapps},
AlgoVeri~\citep{zhao2026algoveri}, DafnyBench~\citep{loughridge2024dafnybench}, the multi-language
vericoding benchmark~\citep{bursuc2025vericoding}, and competitive-programming variants~\citep{xie2026vericontest};
VeriScale~\citep{bai2026veriscale} stresses the \emph{specification} side with adversarial test suites. These
benchmarks evaluate verification on the original task or implementation; RefactorProof instead re-checks the same proof
after the implementation is refactored.

\paragraph{Proof engineering, neural provers, and repair.}
Proof evolution is an established concern in proof engineering. \citet{ringer2019qed} survey proof
maintenance and automation as central challenges in large formal developments. PUMPKIN PATCH
\citep{ringer2018adapting} learns patches from related changes to Coq specifications and proofs, while
PUMPKIN Pi transports proofs across equivalent type changes \citep{ringer2021proofrepair}. Modern
prover-specialized models include Goedel-Prover~\citep{lin2025goedel,lin2025goedelv2} and
DeepSeek-Prover-V2~\citep{ren2025deepseekproverv2}. COPRA uses proof-environment feedback with
backtracking search~\citep{thakur2023copra}, while Lean Copilot integrates tactic suggestion, proof search,
and premise selection~\citep{song2024leancopilot}. Baldur~\citep{first2023baldur} repairs whole proofs from
compiler feedback, and BlueprintRepair~\citep{khrulev2026blueprintrepair} repairs Lean proof blueprints with
typed local edits.
These systems generate, search, or repair proofs after failure. RefactorProof instead freezes the proof and
changes only the implementation, so we measure unchanged-proof survival before any repair is attempted.

\paragraph{Robustness to semantics-preserving perturbations.}
ReCode~\citep{wang2022recode} evaluates code generators under semantics-preserving rewrites of prompt and
program; recent work perturbs the natural-language input of Lean~4 autoformalization~\citep{gui2026robustness}.
Mariposa~\citep{zhou2023mariposa} measures SMT proof instability under semantically irrelevant changes to
verification queries. RefactorProof instead changes the program implementation while holding a checked Lean
proof and its specification fixed. Each retained program change also has a separate Lean equivalence
certificate.

\section{Method}\label{sec:method}
\subsection{Setting and invariants}
A VERINA task file contains marker-delimited regions for imports, auxiliary definitions, the
precondition, the implementation body (\texttt{code}), the postcondition, and the proof of
\texttt{f\_postcond\_satisfied}.  A transformation rewrites only the \texttt{code} region (and, for
helper extraction, appends to the implementation-helper region).  Six invariants are enforced
mechanically: (A) the specification regions are unchanged (SHA-256 asserted); (B) the evaluated proof is
byte-identical to the proof that verified on the original (SHA-256); (C) behavior preservation is certified
by Lean in a separate file; (D) that certificate is never imported by, or textually present in, the
survival file; (E) no \texttt{sorry}, \texttt{admit} or new axioms anywhere; (F) the public function name
and signature are unchanged. The certificate file and survival file are rendered from the same variant
record, and the survival file's specification and proof hashes are checked against the hashes stored in
that record before compilation. For (E), we also ask Lean directly. We append \texttt{\#print axioms} to
copies of all 458 certified equivalence files and all 423 surviving proof files in the released evaluation
and re-check them. None depends on \texttt{sorryAx} or on an axiom beyond \texttt{propext},
\texttt{Classical.choice}, and \texttt{Quot.sound}; 223 depend on no axioms. Two VERINA tasks declare upstream axioms about
\texttt{Float}, but no audited certificate or surviving proof depends on them.

\subsection{Transformation classes}\label{sec:transformations}
\begin{description}\setlength\itemsep{0pt}
  \item[T1 whole-body \texttt{let} introduction ($D$).]  \texttt{def f xs : T := E} becomes
  \texttt{let tmp : T := E; tmp}: definitionally equal by $\zeta$-reduction, a term-level no-op.
  \item[T2 helper extraction ($S$).] The body moves verbatim into a fresh \texttt{private def} and
  \texttt{f} becomes a call to it. The helper remains definitionally reducible, but \texttt{unfold f} and
  \texttt{simp [f]} expose the helper application rather than placing the helper equation in the simplifier.
  \item[T3 commutative operand swap ($E$).]  One occurrence of $A \mathbin{\mathrm{op}} B$ with
  $\mathrm{op} \in \{+,\ast,\texttt{\&\&},\texttt{||},\land,\lor\}$ and atomic operands becomes
  $B \mathbin{\mathrm{op}} A$ (precedence-checked); certified with commutativity/associativity lemmas
  (\texttt{Nat.add\_comm}, \ldots), no type assumed.
  \item[T4 conditional inversion ($E$).]  The tail-position \texttt{if c then A else B} becomes
  \texttt{if ¬c then B else A}; certified via \texttt{ite\_not} or case analysis on \texttt{c}.
  \item[T5 branch extraction ($S$).]  When the body \emph{is} a tail \texttt{if}, its \texttt{else}
  branch moves into a fresh \texttt{private def} with the same binders; delta-equivalent like T2 but
  partial, since the conditional stays and only one branch is extracted.
\end{description}
The $D/S/E$ labels refer to source-level transformation mechanisms. In particular, T2 and T5 are
structural source refactorings even though their equivalence certificates can close by definitional
reduction in Lean.
All transformations are deterministic text rewrites given $(\text{task}, \text{seed})$; no model is
involved. Transformation applicability and rewrite locations are determined from the implementation region
alone; neither the specification nor any proof artifact is inspected when constructing a candidate.
Applicability rules (e.g.\ no def-level \texttt{termination\_by}, no \texttt{where} clause) and every
inapplicable or failed candidate are logged (Table~\ref{tab:construction}).

\subsection{Certification}
For each candidate we emit a file that defines the original and transformed implementations in two
namespaces (sharing the unchanged helpers) and states pointwise equality.  A fixed sequence of independent certificate tactics is attempted in one compile: plain \texttt{rfl}; \texttt{rfl} with smart unfolding disabled
(needed for structurally recursive bodies applied to variables); \texttt{delta} of the definitions and
their \texttt{let rec} auxiliaries, then \texttt{rfl}; \texttt{simp only} and \texttt{simp} with the
family's laws; and family-specific tactics (\texttt{ac\_nf}, \texttt{by\_cases}).  A variant is certified
if any level succeeds and \emph{rfl-certified} if one of the first three (definitional) levels does.  Where
Lean~4.18's code generator refuses a wrapped body, the definition is marked \texttt{noncomputable} (name and
signature unchanged) and the retry is recorded.

\subsection{Survival evaluation and metrics}
For each certified variant and each proof artifact we rebuild the task file with only the implementation
regions changed and compile it in the pinned VERINA environment (Lean~v4.18.0~\citep{demoura2021lean4},
Mathlib~\citep{mathlib2020} pinned by \texttt{lake-manifest.json}).  Errors are attributed to the proof region; failures elsewhere are excluded as
build errors (none occurred).

\paragraph{Definitions.}  For a task $i$ with a fixed proof artifact $P_i$ and certified variants
$V_i = \{C'_{i1},\dots,C'_{im_i}\}$ of its implementation $C_i$, where every variant satisfies the
Lean-checked certificate
\begin{equation}
  \forall \bar{x}.\; C_i(\bar{x}) = C'_{ij}(\bar{x}) \qquad\text{(proved in a separate file, never imported)},
\label{eq:cert}
\end{equation}
the per-task survival rate and the task-normalized aggregate over the $N$ tasks that have a valid proof
are
\begin{equation}
  \mathrm{PSR}_i = \frac{1}{m_i}\sum_{j=1}^{m_i} \mathbb{1}\big[\,P_i \text{ verifies against } (C'_{ij}, S_i)\,\big],
  \qquad
  \mathrm{PSR}_{\text{task}} = \frac{1}{N}\sum_{i=1}^{N} \mathrm{PSR}_i ,
\label{eq:psr}
\end{equation}
with 95\% intervals from a cluster bootstrap over tasks (10{,}000 replicates). We report
$\mathrm{PSR}_{\text{task}}$ overall and within each transformation class $\sigma \in \{D,S,E\}$.
For a class-specific PSR, variants in that class are first averaged within task, and the resulting task
rates are then averaged over tasks that have at least one certified variant in the class.
RefactorProof construction uses the 185 tasks whose original files compile in the pinned environment;
model coverage uses all 189 VERINA tasks attempted by the generation protocol. For a model with $K$
samples per task in fixed order, coverage is
\begin{equation}
  \mathrm{Coverage}@k = \frac{1}{|T|}\sum_{i\in T} \mathbb{1}\big[\exists\, s \le k:\ \text{sample } s
  \text{ of task } i \text{ verifies on } (C_i,S_i)\big], \qquad k = 1,\dots,K,
\label{eq:coverage}
\end{equation}
over all $|T|=189$ attempted tasks; the primary artifact is the first verifying sample.  Two proof
sources $a, b$ are compared on the matched task set $T_{ab}$ where both have a valid proof, by the
paired difference
\begin{equation}
  \Delta_i = \mathrm{PSR}_i^{(a)} - \mathrm{PSR}_i^{(b)}, \qquad
  \overline{\Delta} = \frac{1}{|T_{ab}|}\sum_{i \in T_{ab}} \Delta_i ,
\label{eq:paired}
\end{equation}
with a task-cluster bootstrap interval. Sign-flip permutation tests are also computed and released,
but the main text emphasizes effect sizes and confidence intervals because the class-specific comparisons
are exploratory and are not multiplicity-corrected. Both sources are evaluated on identical variants, so
$\Delta_i$ isolates the proof.

\subsection{Evaluation populations and reference-proof availability}\label{sec:populations}
VERINA contains 189 tasks, but ground-truth proofs are optional in the source benchmark. The VERINA
authors manually provide reference proofs for 46 tasks for quality assurance \citep{ye2025verina}. They are
correctness witnesses for the upstream benchmark, not proofs written or reviewed for robustness under
program change, and their style reflects that purpose: 45 of the 46 open by unfolding the target
definition. All 46 are in VERINA's basic split. Their median implementation length is one line and 6.5\% are recursive. Among
the other 139 tasks in the RefactorProof construction population, the median is eight lines, 54.0\% are
recursive, and 56.1\% are in the advanced split. Reference-proof PSR therefore describes a simpler upstream
subset rather than the full benchmark. These 46 tasks contribute 130 certified variants. RefactorProof
construction instead uses the 185 original task files that compile in our pinned environment. Model
coverage is measured over all 189 attempted tasks, and model PSR is conditional on the model first producing
a Lean-valid proof of the original program. Table~\ref{tab:populations} makes these denominators explicit.

\begin{table}[t]\centering\small
\caption{Task populations and original-proof coverage. Reference proofs are supplied upstream. Model
Coverage@$k$ is the fraction of all 189 tasks with a Lean-valid proof within the first $k$ samples; ``valid
artifacts'' counts all valid samples among $K=5$.}\label{tab:populations}
\resizebox{\linewidth}{!}{\begin{tabular}{lrrrr}
\toprule
Population / proof source & Tasks & Coverage@1 (\%) & Coverage@5 (\%) & Valid artifacts \\
\midrule
VERINA source benchmark & 189 & -- & -- & -- \\
RefactorProof construction & 185 & -- & -- & -- \\
VERINA reference proofs & 46 & -- & -- & 46 supplied \\
DeepSeek-Prover-V2-7B & 31/189 & 12.7 & 16.4 & 111 \\
Goedel-Prover-V2-32B & 26/189 & 11.6 & 13.8 & 108 \\
Goedel-Prover-V2-8B & 29/189 & 9.5 & 15.3 & 100 \\
Qwen3-14B & 5/189 & 0.5 & 2.6 & 8 \\
Qwen3-32B & 5/189 & 1.6 & 2.6 & 7 \\
\bottomrule
\end{tabular}
}
\end{table}

\subsection{LLM proofs}
Proofs are generated on the original implementations only with VERINA's prompting code. We draw $K=5$
independent samples per task in fixed order and use the first Lean-valid sample as the primary proof.
The evaluated checkpoints are \texttt{Qwen/Qwen3-14B}, \texttt{Qwen/Qwen3-32B}~\citep{yang2025qwen3},
\texttt{Goedel-LM/Goedel-Prover-V2-8B}, \texttt{Goedel-LM/Goedel-Prover-V2-32B}, and
\texttt{deepseek-ai/DeepSeek-Prover-V2-7B}. General models use VERINA's proof-generation prompt; the
specialized provers use VERINA's prover template. All requests use temperature 1.0 and a 16{,}000-token
output limit. Qwen3-14B, Qwen3-32B, and Goedel-8B inherit model defaults of top-$p=0.95$ and top-$k=20$;
Goedel-32B and DeepSeek-7B use top-$p=1.0$ with no top-$k$ truncation. Each model is served with vLLM~\citep{kwon2023vllm} on one
NVIDIA A100-80GB. The two 32B models use FP8 weight-only quantization; the others use their bf16 checkpoints.
Individual generation runs require 1.5 to 9 GPU-hours. Lean checking runs on 24-core CPU nodes: the
dedicated Lean jobs consume 24.9 CPU-hours of CPU time out of 39.9 allocated, and further Lean checking runs
inside the generation jobs, whose 146.1 CPU-hours cover generation and checking together. Counting every
scheduled job, including failed and exploratory runs, the project used 35.3 GPU-hours. Lean validation uses v4.18.0
with VERINA's pinned Mathlib and a 300-second timeout per file.

To test dependence on the first-valid rule, we re-check all 334 Lean-valid samples on every certified
variant and recompute matched comparisons using the first valid proof, the shortest valid proof, a fixed-seed
random valid proof, and the average over all valid proofs. We also recompute PSR after excluding T1, with
family-balanced and one-per-task-family weighting, and on common task-family support across proof sources.

\begin{figure}[H]
\centering
\includegraphics[width=\linewidth]{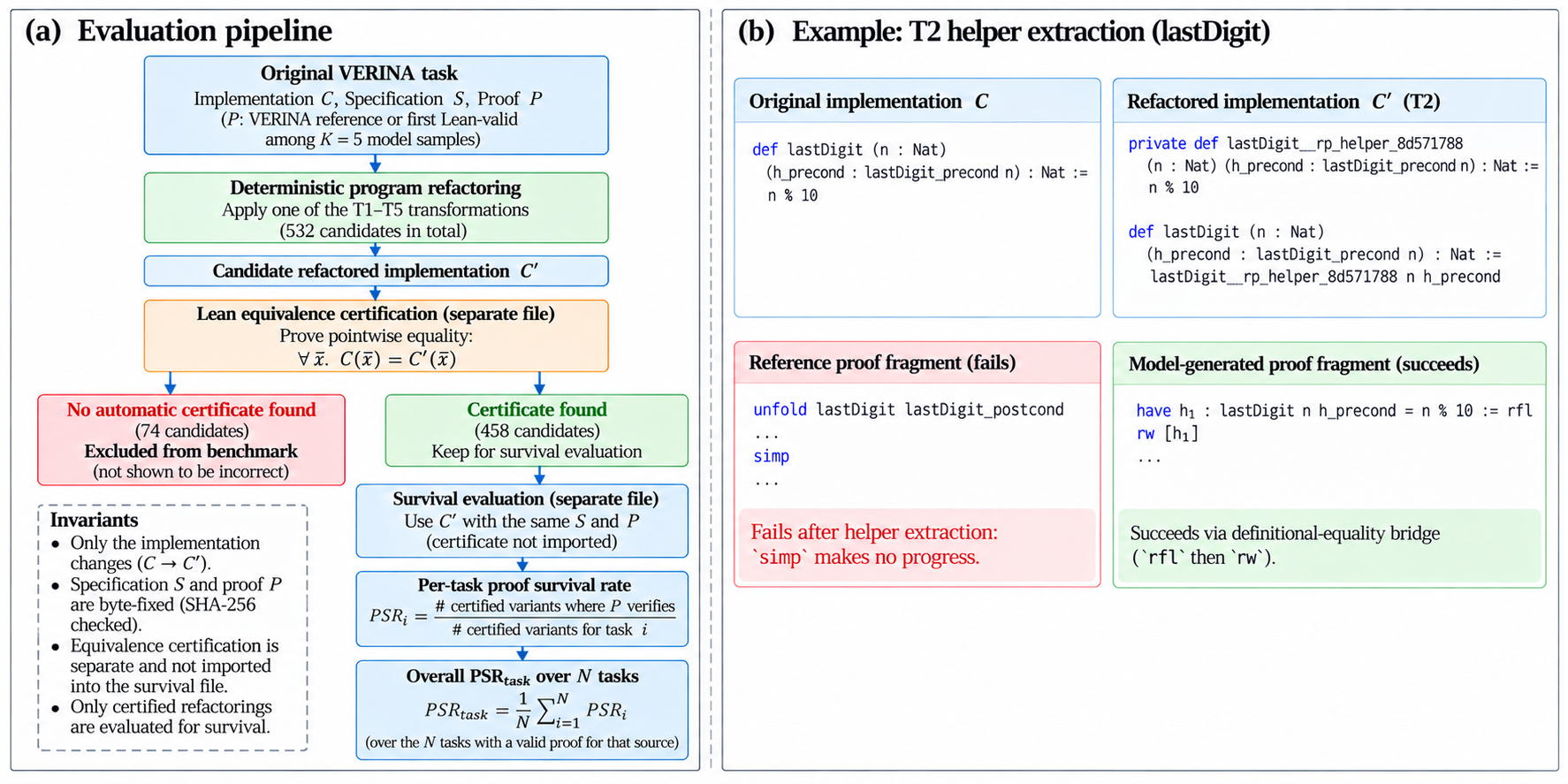}
\caption{RefactorProof evaluation design and a T2 helper-extraction example. (a) Each candidate refactoring
$C'$ is kept only if Lean separately proves $\forall \bar{x}.\,C(\bar{x})=C'(\bar{x})$; candidates for which
the fixed automatic certificate search finds no proof are excluded and are not thereby shown incorrect.
The specification $S$ and proof $P$ remain byte-identical, the certificate is never imported into the
survival file, and $\mathrm{PSR}_{\mathrm{task}}$ is averaged over tasks with a valid proof for that source.
(b) VERINA task \texttt{lastDigit}. The supplied reference proof fails after T2 helper extraction with
\texttt{simp made no progress}; the two model lines shown are verbatim from the DeepSeek-Prover-V2-7B
primary proof and survive via definitional equality.}
\label{fig:overview}
\end{figure}

Figure~\ref{fig:overview} summarizes the certification and survival pipeline and gives a concrete T2
helper-extraction example.

\section{Results}\label{sec:results}
\subsection{RefactorProof construction}
Certification succeeds for nearly all T1, T2, T4, and T5 candidates (Table~\ref{tab:construction}). T3 is
different: only \certTthree{} candidates certify, and the rejected cases are overwhelmingly recursive
(98.6\%, compared with 14.0\% among certified candidates). The fixed certificate search does not synthesize
the induction often needed for recursive swaps. T3 results therefore describe a selected, mostly
non-recursive subset; additional construction diagnostics are reported in Appendix~\ref{sec:sensitivity}.

\begin{table}[H]\centering\small
\caption{RefactorProof construction over the \nCompiling{} VERINA tasks whose originals compile in the pinned
environment. Certification rate is over generated candidates; ``rfl'' is the share of certified variants
closed at a definitional level.}\label{tab:construction}
\resizebox{\linewidth}{!}{\begin{tabular}{llrrrrrr}
\toprule
Transformation & Class & Applicable tasks & Candidates & Certified & Cert. rate (\%) & rfl (\%) & Tasks w/ variant \\
\midrule
T1\_let\_intro & D & 181/185 & 181 & 180 & 99.4 & 100.0 & 180 \\
T2\_helper\_extract & S & 179/185 & 179 & 178 & 99.4 & 100.0 & 178 \\
T3\_comm\_swap & E & 77/185 & 122 & 50 & 41.0 & 4.0 & 34 \\
T4\_cond\_invert & E & 32/185 & 32 & 32 & 100.0 & 0.0 & 32 \\
T5\_branch\_extract & S & 18/185 & 18 & 18 & 100.0 & 100.0 & 18 \\
\bottomrule
\end{tabular}
}
\end{table}

\subsection{Proof stability by source}
Coverage and PSR answer different questions. Coverage@5 is only 13.8--16.4\% for the three specialized
provers (Table~\ref{tab:populations}), so their PSR estimates are conditional on first producing a valid proof
of the original task. On the matched supplied-reference subset, however, all three specialized provers have
higher overall PSR than the reference proofs: the differences range from +15.6 to \deltaGoedelAll{} percentage
points, and all three overall 95\% bootstrap intervals exclude zero (Table~\ref{tab:matched}). We therefore use
the overall matched comparison as the primary source-level result; the family-specific rows are exploratory.

\begin{table}[H]\centering\small
\caption{Task-normalized PSR for the reference proofs and specialized provers. Coverage columns report
Coverage@1 and Coverage@5 over all 189 attempted tasks; reference proofs are supplied rather than generated.
T3 can contribute more than one variant per task, so each reported group is first averaged within task.}\label{tab:main}
\resizebox{\linewidth}{!}{\begin{tabular}{lrrrlrrr}
\toprule
\multicolumn{8}{c}{\textbf{Panel A: Overall and transformation class}} \\
\cmidrule(lr){1-8}
Proof source & C@1 & C@5 & Tasks & Overall PSR (\%) [95\% CI] & D & S & E \\
\midrule
DeepSeek-Prover-V2-7B & 12.7 & 16.4 & 31 & 59.3 [52.4, 66.4] & 100.0 & 13.3 & 72.2 \\
Goedel-Prover-V2-32B & 11.6 & 13.8 & 26 & 64.5 [56.5, 72.8] & 100.0 & 26.0 & 67.6 \\
Goedel-Prover-V2-8B & 9.5 & 15.3 & 29 & 64.6 [57.8, 72.1] & 100.0 & 26.8 & 72.4 \\
reference & -- & -- & 46 & 47.9 [44.0, 52.0] & 100.0 & 4.3 & 44.2 \\
\bottomrule
\end{tabular}

\vspace{3pt}

\begin{tabular}{lrrrrr}
\toprule
\multicolumn{6}{c}{\textbf{Panel B: Transformation family}} \\
\cmidrule(lr){1-6}
Proof source & T1 & T2 & T3 & T4 & T5 \\
\midrule
DeepSeek-Prover-V2-7B & 100.0 & 10.0 & 54.5 & 100.0 & 22.2 \\
Goedel-Prover-V2-32B & 100.0 & 24.0 & 57.6 & 87.5 & 25.0 \\
Goedel-Prover-V2-8B & 100.0 & 25.0 & 60.6 & 90.0 & 20.0 \\
reference & 100.0 & 2.2 & 18.2 & 66.7 & 18.2 \\
\bottomrule
\end{tabular}
}
\end{table}

Coverage changes the end-to-end interpretation. Over the 180 tasks with at least one certified variant, the
joint rate that a valid proof artifact is available for a source and survives a uniformly sampled certified
variant is 12.2\% for the supplied reference proofs, compared with 9.3--10.4\% for the three specialized
provers. For the reference source, availability means that VERINA supplies a proof; for a model, it means that
the model produces a Lean-valid proof. Thus, the provers' higher conditional PSR does not yield a higher
artifact-availability-and-survival rate. The Qwen controls cover only five tasks each, so we do not draw
source-level robustness conclusions from them.

\begin{table}[H]\centering\small
\caption{Matched model versus supplied-reference comparison on identical certified variants. Values are
paired PSR differences with task-cluster bootstrap 95\% CIs. Overall sensitivity checks remain positive:
excluding T1 gives +24.3, +32.9, and +31.3 points for DeepSeek-7B, Goedel-8B, and Goedel-32B;
family-balanced aggregation gives +13.7, +18.6, and +17.3 points. PSR columns are rounded independently;
$\Delta$PSR is computed from the unrounded task-level values. Appendix~\ref{sec:sensitivity} reports full
sensitivity results.}\label{tab:matched}
\resizebox{\linewidth}{!}{\begin{tabular}{llrrrrr}
\toprule
Model & Class & Matched tasks & Ref. PSR & Model PSR & $\Delta$PSR & 95\% CI \\
\midrule
deepseek-prover-v2-7b & all & 25 & 43.6 & 59.2 & +15.6 & [8.0, 24.3] \\
deepseek-prover-v2-7b & structural & 25 & 2.0 & 16.0 & +14.0 & [2.0, 30.0] \\
deepseek-prover-v2-7b & semantic & 15 & 35.6 & 73.3 & +37.8 & [13.3, 62.2] \\
\midrule
goedel-prover-v2-32b & all & 21 & 43.5 & 62.4 & +18.8 & [10.1, 28.1] \\
goedel-prover-v2-32b & structural & 21 & 2.4 & 26.2 & +23.8 & [4.8, 45.2] \\
goedel-prover-v2-32b & semantic & 14 & 38.1 & 67.9 & +29.8 & [0.0, 58.3] \\
\midrule
goedel-prover-v2-8b & all & 23 & 43.0 & 63.4 & +20.3 & [12.0, 29.0] \\
goedel-prover-v2-8b & structural & 23 & 2.2 & 23.9 & +21.7 & [4.3, 39.1] \\
goedel-prover-v2-8b & semantic & 15 & 35.6 & 78.3 & +42.8 & [11.1, 71.1] \\
\bottomrule
\end{tabular}
}
\end{table}

The overall matched ordering is stable under alternative valid-proof selection rules, exclusion of T1,
family-balanced aggregation, and common task-family support (Appendix~\ref{sec:sensitivity}). The
class-specific intervals are less stable: the Goedel-32B semantic interval reaches zero, and the DeepSeek
structural lower bound varies from 0.0 to 2.0 across bootstrap seeds, so we treat family-level rows as
exploratory. Taskwise win/loss/tie counts show that the overall differences are not driven by a few
reference-favoring reversals. T4 is the clearest family-level case, while T3 and T5 have only 7--9 matched
tasks. Full sensitivity and proof-length checks are reported in Appendix~\ref{sec:sensitivity}.

\subsection{Why proof strategies respond differently to refactoring}
Reference proofs usually begin with \texttt{unfold}, whereas the specialized provers more often name the
implementation in \texttt{simp} or \texttt{dsimp} and continue with automation (Table~\ref{tab:profile}).
This stylistic difference is not itself a causal explanation. Inspection of compiled survivors and failures
instead points to one specific mechanism: an explicit definitional bridge of the form
\texttt{have h : f args = body := rfl}.

\begin{table}[H]\centering\small
\caption{Tactic profile of the fixed reference and specialized-prover proof artifacts.}\label{tab:profile}
\resizebox{.92\linewidth}{!}{\begin{tabular}{lrrrrr}
\toprule
Proof source & $n$ & unfold target (\%) & simp[target] (\%) & med. LoC & med. tactics \\
\midrule
deepseek-prover-v2-7b & 31 & 9.7 & 77.4 & 25 & 23 \\
goedel-prover-v2-32b & 26 & 0.0 & 84.6 & 36 & 33 \\
goedel-prover-v2-8b & 29 & 0.0 & 79.3 & 32 & 30 \\
reference & 46 & 97.8 & 2.2 & 11 & 12 \\
\bottomrule
\end{tabular}
}
\end{table}

Under structural helper extraction, such bridges are associated with much higher survival. At task level,
bridge proofs survive in 6/6 tasks where they occur, compared with 6/31 tasks for other strategies; on the
same refactored variants, the corresponding counts are 9/9 and 2/9 (Table~\ref{tab:mechanism}). Figure~\ref{fig:overview}
illustrates the mechanism: the bridge rewrites the function call to its body using definitional equality,
whereas later syntax-specific steps can stall after a helper is introduced.

\begin{table}[H]\centering\small
\caption{Mechanism checks that reduce dependence among repeated records. Task-level rows count tasks with at
least one survival in the indicated category. Same-variant rows compare proof categories on the same
refactored variants.}\label{tab:mechanism}
\resizebox{.96\linewidth}{!}{\begin{tabular}{lllll}
\toprule
Setting & Comparison level & Group A & Group B & Evidence \\
\midrule
Structural & task & rfl bridge: 6/6 & other: 6/31 & Fisher $p=0.0004$ \\
Structural & same variant & rfl bridge: 9/9 & other: 2/9 & 6 variants, 5 tasks \\
Semantic & task & rfl bridge: 0/3 & simp[f]: 17/19 & Fisher $p=0.0065$ \\
Semantic & same variant & rfl bridge: 0/4 & simp[f]: 3/5 & 3 variants \\
\bottomrule
\end{tabular}
}
\end{table}

The same bridge is not uniformly stable. Semantic refactorings change the literal body, so an \texttt{rfl}
equation tied to the original body can stop typechecking, while normalization-based proofs can still absorb
the rewrite. At task level, \texttt{rfl}-bridge proofs survive 0/3 semantic cases, compared with 17/19 for
\texttt{simp [f]}-style proofs; on the same semantic variants the counts are 0/4 and 3/5
(Table~\ref{tab:mechanism}). Stability therefore depends on how a proof is coupled to the implementation,
not on one uniformly robust tactic. A prompt-time intervention does not produce a consistent causal effect of
forcing one tactic, so we treat these mechanism results as descriptive associations; record-level breakdowns
and intervention details are in Appendix~\ref{sec:sensitivity}.

\subsection{Mechanical repair of structural failures}\label{sec:repair}
PSR asks whether a proof survives unchanged, not how difficult it is to repair after failure. We therefore
predefine three one-identifier edits that add the extracted helper name to an existing \texttt{unfold},
\texttt{simp}, or \texttt{dsimp} call and recompile the edited proof on the same variant. An unedited control
reproduces all 139 attempted failures. The predefined edit repairs 109/139 eligible failures, including 51/54
reference-proof failures and 58/85 prover failures, with a median edit of two tokens on one line. Many
structural failures are therefore cheap under this narrow repair template even when the original proof does
not survive unchanged. The source- and template-level breakdown is reported in Appendix~\ref{sec:sensitivity}.

\section{Limitations}\label{sec:threats}
\paragraph{Design-time versus post hoc choices.}
The five transformation families, the $K=5$ sampling protocol with first-valid selection, the
task-normalized PSR, the $D/S/E$ ladder, and the bootstrap settings were fixed in a frozen configuration
before any proof was generated. The mechanism categories, repair templates, alternative weightings, and the
selection-rule and bootstrap-seed sensitivity analyses were specified after the survival outcomes were
known. We report these later analyses as exploratory. For the repair experiment specifically, the three
templates were fixed before any repair outcomes were evaluated.

\paragraph{Certification selection.}
Only certified RefactorProof variants enter PSR. T3 is strongly selected by certificate strength: 14.0\% of
certified candidates are recursive versus 98.6\% of rejected candidates. The automatic sequence does not
synthesize the induction needed for many recursive swaps. T3 results therefore apply to the certified,
mostly non-recursive subset rather than to all behavior-preserving commutative rewrites.

\paragraph{Reference population, coverage, and proof selection.}
The reference arm is limited to 46 basic-split VERINA tasks and is substantially simpler than the rest of
the construction population. The primary matched comparisons further narrow to 21--25 short,
non-recursive basic-split tasks; source-level differences therefore characterize proof artifacts on this
slice, not checkpoint-wide robustness over VERINA. Model PSR is also conditional on solving the original
task, with Coverage@5 between 13.8\% and 16.4\% for the three specialized provers. Alternative deployable selection rules preserve every overall matched advantage,
but a post hoc least-robust-valid bound can remove the structural advantage. Structural robustness therefore
depends in part on which valid proof is selected. The specialized provers evaluated here are open-weight
models no larger than 32B parameters, and the two Qwen controls cover only five tasks each.

\paragraph{Benchmark exposure.}
VERINA is publicly available and predates the checkpoints we evaluate, so we cannot rule out that these
tasks, or proofs of them, appeared in pretraining. Such exposure could inflate original-task coverage. Its
effect on survival is less direct because the refactored implementations and their certificates are generated
here and were not public before this work.

\paragraph{Prompt and strategy confounds.}
Proof length does not explain the source gap in our regression or overlapping-length analysis, but the source
comparison remains observational. The general-model and prover templates differ, model-specific decoding
defaults differ, and the 32B runs use FP8 weight-only quantization. The within-prover intervention also fails
to establish a common causal effect of one tactic. We therefore treat the between-source tactic profile as
descriptive and ground mechanism claims in compiled survivor, failure, task-level, and same-variant checks.

\paragraph{Statistics and mechanism sample size.}
Class- and family-specific comparisons are exploratory and are not multiplicity-corrected. T3 and T5 have
only 7--9 matched tasks. Two class-level percentile-bootstrap intervals also touch zero under seed
sensitivity, so we do not describe them as excluding zero. Mechanism support is small: the structural
\texttt{rfl}-bridge pattern spans six tasks and the semantic pattern three. The categories were developed
post hoc from the observed artifacts. Task-level Fisher tests, leave-one-task-out checks, and same-variant
comparisons reduce dependence concerns but do not turn these associations into causal tactic effects.

\paragraph{Transformation scope and practical maintenance.}
RefactorProof uses one pinned Lean/Mathlib environment and five local, mechanical transformation families.
They do not cover algorithm changes, data-representation changes, argument reordering, inlining, or the full
distribution of refactorings encountered in software projects. Four VERINA tasks do not compile in the
pinned environment. Two certified variants require \texttt{noncomputable} to work around a Lean code-generation
limitation. Their pointwise equivalence certificates remain kernel-checked, but the transformed definitions
are not compiled by Lean's code generator, so this workaround does not preserve ordinary executability.
Excluding these two variants changes no reported statistic by more than 0.2 points. We retain them in the
primary benchmark because inclusion is defined by kernel-checked pointwise equivalence rather than code-generation
eligibility; the exclusion analysis quantifies the effect of adopting the stricter executable-only criterion.
The repair experiment covers only failures eligible for three predefined \texttt{unfold}/\texttt{simp}/\texttt{dsimp} edits. It is not a re-proving experiment and does
not estimate general maintenance cost after arbitrary refactorings. The T2 and T5 results also depend on
Lean's transparency settings: the extracted helper is a \texttt{private def} that remains definitionally
reducible, which allows reflexivity-style bridges to survive. Marking such a helper \texttt{@[irreducible]}
or changing unfolding behavior through \texttt{set\_option} could change which proofs survive; we did not
vary this.

\section{Conclusion}
RefactorProof tests whether an already valid Lean proof remains valid after a certified behavior-preserving
change to the implementation. On the short, non-recursive matched VERINA tasks with supplied reference proofs, the three
specialized provers have higher overall unchanged-proof survival, although family-specific evidence is
strongest for conditional inversion and helper extraction. The inspected artifacts remain coupled to
implementation details through different mechanisms: syntax-specific rewrites, definitional-equality
bridges, and normalization fail under different changes. Many eligible structural failures are repaired by
a one-identifier edit, so PSR should be read as proof-artifact stability rather than maintenance cost.
One-time verification does not establish that the same proof will remain reusable unchanged after refactoring.

\begin{ack}
The authors received no specific funding for this work.
\end{ack}

\bibliographystyle{plainnat}
\bibliography{refs}

\newpage
\appendix
\section{Additional sensitivity analyses}\label{sec:sensitivity}
This appendix reports the additional analyses used to test the main comparisons and their scope. All values
come from the same frozen generation outputs or from Lean re-checks in the pinned VERINA environment; no new
model proof is generated for these analyses.

\subsection{Valid-proof selection}
All 334 Lean-valid samples are re-checked on every certified variant. The first-valid re-evaluation reproduces
301/301 frozen primary survival outcomes. Table~\ref{tab:selection_sensitivity} reports overall matched
$\Delta$PSR under four deployable summaries. Every overall 95\% task-cluster bootstrap interval excludes
zero under each rule. Class-specific conclusions are less stable: the DeepSeek structural interval touches
zero under first-valid selection and crosses zero under shortest-valid selection; the Goedel-32B semantic
interval also reaches or crosses zero under these rules.

\begin{table}[h]\centering\small
\caption{Overall matched $\Delta$PSR (percentage points) under valid-proof selection rules. ``All valid''
averages each task across all valid samples.}\label{tab:selection_sensitivity}
\begin{tabular}{lrrrr}
\toprule
Model & First valid & Shortest valid & Random valid & All valid \\
\midrule
DeepSeek-Prover-V2-7B & +15.6 & +11.3 & +15.6 & +15.5 \\
Goedel-Prover-V2-8B & +20.3 & +22.4 & +20.7 & +20.7 \\
Goedel-Prover-V2-32B & +18.8 & +24.5 & +20.9 & +20.9 \\
\bottomrule
\end{tabular}
\end{table}

A post hoc least-robust-valid bound is deliberately not treated as a deployable selection rule. Under that
bound, structural $\Delta$PSR is +6.5 [-4.3, 19.6] for Goedel-8B, +2.4 [-7.1, 14.3] for Goedel-32B, and
-2.0 [-6.0, 0.0] for DeepSeek-7B. This shows that structural robustness can depend on which valid proof a
model happens to emit.

\subsection{Within-task variability across valid proofs}\label{sec:within_task_variability}
The proof-selection analysis re-checks every Lean-valid sample on every certified variant of its task, which
also lets us measure variation among valid proofs of the same model--task pair. We summarize the three
specialized provers because the Qwen controls have only two tasks each with multiple valid proofs.

\begin{table}[H]\centering\scriptsize
\caption{Within-task variability across independently sampled Lean-valid proofs. ``Identical'' is the share
of tasks on which all valid samples have the same survive/fail vector over certified variants. Ranges are
mean within-task max--min PSR differences in percentage points; $S$ and $E$ restrict to structural and
semantic variants.}
\label{tab:within_task_variability}
\resizebox{\linewidth}{!}{%
\begin{tabular}{lrrrrrrr}
\toprule
Model & Tasks ($\geq2$ valid) & Identical (\%) & PSR range & $S$ range & $E$ range & Within-var. share (\%) & $|$first--mean$|$ \\
\midrule
DeepSeek-7B & 26 & 46 & 17.4 & 48.0 & 35.3 & 41 & 6.7 \\
Goedel-8B & 25 & 44 & 15.1 & 36.0 & 45.4 & 29 & 4.7 \\
Goedel-32B & 24 & 29 & 22.4 & 54.2 & 55.9 & 54 & 8.6 \\
\bottomrule
\end{tabular}%
}
\end{table}

Different valid proofs disagree on at least one variant for 54\%, 56\%, and 71\% of the multi-proof tasks
for DeepSeek-7B, Goedel-8B, and Goedel-32B, respectively. Mean within-task PSR ranges are 15.1--22.4
points, and within-task variance accounts for 29--54\% of the combined within- and between-task variance.
Structural survival is especially sample-dependent, with mean within-task ranges of 36--54 points.
Survival therefore varies materially across valid proof artifacts from the same model and task. However,
the first-valid artifact differs from the within-task mean by only 4.7--8.6 points on average, and every
deployable selection rule we test preserves the direction of all three overall matched differences
(Table~\ref{tab:selection_sensitivity}). Thus, proof choice contributes substantial within-task variation, but the
primary overall matched ordering is not an artifact of the first-valid selection rule. Because most tasks
have only a few certified variants, these ranges are coarse; the identical-outcome share is the cleaner
summary.

\subsection{Aggregation and common support}
The end-to-end quantity combines proof-artifact availability with survival on a uniformly sampled certified
variant. For the reference source, availability means that VERINA supplies a proof; for a model, it means
that the model produces a Lean-valid proof. Table~\ref{tab:end_to_end} reports the estimates and their
uncertainty.

\begin{table}[H]\centering\small
\caption{Artifact-availability-and-survival rate over the 180 tasks with at least one certified variant.}
\label{tab:end_to_end}
\begin{tabular}{lcc}
\toprule
Proof source & Rate (\%) & 95\% CI \\
\midrule
VERINA supplied reference & 12.2 & [9.0, 15.5] \\
DeepSeek-Prover-V2-7B & 10.2 & [6.8, 13.8] \\
Goedel-Prover-V2-8B & 10.4 & [6.9, 14.2] \\
Goedel-Prover-V2-32B & 9.3 & [5.9, 13.0] \\
\bottomrule
\end{tabular}
\end{table}

T1 and T2 contribute 358/458 certified variants, so we also vary the aggregation and support used for the
matched comparison. Table~\ref{tab:aggregation_sensitivity} collects the corresponding model-minus-reference
$\Delta$PSR values. Excluding T1, source-level PSR is 11.5\% for reference proofs, 32.0\% for DeepSeek-7B,
43.8\% for Goedel-8B, and 43.7\% for Goedel-32B. Family-balanced and one-per-task-family weighting move any
source's PSR by at most 3.8 points, and the reference source remains lowest under all tested weightings.

\begin{table}[H]\centering\scriptsize
\caption{Matched $\Delta$PSR sensitivity to aggregation and common support. Values are percentage points with
95\% task-cluster bootstrap intervals. The final row uses the 65 task-family cells shared by all four sources
(21 tasks).}
\label{tab:aggregation_sensitivity}
\begin{tabular}{lccc}
\toprule
Analysis & DeepSeek-7B & Goedel-8B & Goedel-32B \\
\midrule
Exclude T1 & +24.3 [11.5, 38.9] & +32.9 [18.4, 48.6] & +31.3 [15.7, 48.7] \\
Family-balanced within task & +13.7 [6.8, 21.5] & +18.6 [10.9, 26.6] & +17.3 [9.2, 26.2] \\
Exclude T1 and T2 & +28.7 [13.3, 46.0] & +30.7 [4.0, 54.7] & +20.7 [-4.3, 45.7] \\
Common task-family support & +17.4 [8.3, 26.8] & +21.1 [12.3, 30.2] & +18.8 [9.9, 27.8] \\
\bottomrule
\end{tabular}
\end{table}

For the T1-and-T2-excluded comparison, only 14--15 matched tasks remain; sign-flip $p$-values are 0.017,
0.066, and 0.216 for DeepSeek-7B, Goedel-8B, and Goedel-32B, respectively. Thus, on the semantic-and-T5
remainder the point estimate stays positive for all three sources, but the Goedel-32B interval crosses zero.

\paragraph{Taskwise direction of matched effects.}
The aggregate matched differences are not produced by repeated reversals in favor of the reference proofs.
Table~\ref{tab:win_loss_tie} counts tasks according to which source has the higher per-task PSR. In the overall
comparisons, no task favors the reference proof; in the class-specific comparisons, the reference proof leads
on at most one task.

\begin{table}[H]\centering\scriptsize
\caption{Taskwise direction of matched PSR differences. ``Model higher'' and ``Reference higher'' compare the
per-task PSR of the two proof artifacts; the remaining tasks tie.}
\label{tab:win_loss_tie}
\begin{tabular}{llrrrrr}
\toprule
Model & Class & Tasks & Model higher & Reference higher & Tie & $\Delta$PSR \\
\midrule
DeepSeek-7B & all & 25 & 10 & 0 & 15 & +15.6 \\
DeepSeek-7B & structural & 25 & 5 & 1 & 19 & +14.0 \\
DeepSeek-7B & semantic & 15 & 6 & 0 & 9 & +37.8 \\
Goedel-8B & all & 23 & 12 & 0 & 11 & +20.3 \\
Goedel-8B & structural & 23 & 6 & 1 & 16 & +21.7 \\
Goedel-8B & semantic & 15 & 8 & 1 & 6 & +42.8 \\
Goedel-32B & all & 21 & 10 & 0 & 11 & +18.8 \\
Goedel-32B & structural & 21 & 6 & 1 & 14 & +23.8 \\
Goedel-32B & semantic & 14 & 6 & 1 & 7 & +29.8 \\
\bottomrule
\end{tabular}
\end{table}

\subsection{Matched-subset characterization}\label{sec:matched_subset_characterization}
The primary matched comparisons use 21--25 tasks from the 46-task supplied-reference subset. Table~\ref{tab:matched_subset_characterization}
characterizes those tasks relative to the full construction population and the supplied-reference subset.

\begin{table}[H]\centering\scriptsize
\caption{Characteristics of the task populations entering the matched comparisons. T3--T5 columns give the
share of tasks with at least one certified variant of that family. T1 and T2 are available for every matched
task.}
\label{tab:matched_subset_characterization}
\resizebox{\linewidth}{!}{%
\begin{tabular}{lrrrrrrrr}
\toprule
Population & Tasks & Basic (\%) & Recursive (\%) & Med. impl. LoC & Med. postcond. LoC & T3 (\%) & T4 (\%) & T5 (\%) \\
\midrule
Construction population & 185 & 58 & 42.2 & 6 & 3 & 18 & 17 & 10 \\
Supplied-reference subset & 46 & 100 & 6.5 & 1 & 2 & 24 & 26 & 24 \\
Matched, DeepSeek-7B & 25 & 100 & 0.0 & 1 & 1 & 36 & 32 & 32 \\
Matched, Goedel-8B & 23 & 100 & 0.0 & 1 & 1 & 39 & 35 & 35 \\
Matched, Goedel-32B & 21 & 100 & 0.0 & 1 & 1 & 43 & 33 & 33 \\
\bottomrule
\end{tabular}%
}
\end{table}

The matched subsets are a particularly simple slice of VERINA: all are basic-split, none is recursive, and
the median implementation and postcondition are each one line. By comparison, 42.2\% of the construction
population is recursive and its median implementation is six lines. The matched tasks nevertheless have
higher T3--T5 availability than the full population. The model-versus-reference comparison should therefore
be read as evidence for short, non-recursive programs on which several refactoring families are testable; it
does not establish the same source ordering for the benchmark's recursive or longer tasks.

\subsection{Matched effects by transformation family}
\begin{table}[H]\centering\scriptsize
\caption{Matched model-minus-reference $\Delta$PSR by transformation family. Task counts differ because a
family must be present on a task for both sources. T1 is omitted because both sources survive it universally.}
\label{tab:family_matched}
\begin{tabular}{llrrl}
\toprule
Model & Family & Matched tasks & $\Delta$PSR & 95\% CI \\
\midrule
DeepSeek-Prover-V2-7B & T2 & 25 & +12.0 & [0.0, 24.0] \\
Goedel-Prover-V2-8B & T2 & 23 & +21.7 & [4.3, 39.1] \\
Goedel-Prover-V2-32B & T2 & 21 & +23.8 & [9.5, 42.9] \\
DeepSeek-Prover-V2-7B & T3 & 9 & +33.3 & [0.0, 66.7] \\
Goedel-Prover-V2-8B & T3 & 9 & +40.7 & [-3.7, 81.5] \\
Goedel-Prover-V2-32B & T3 & 9 & +25.9 & [-14.8, 66.7] \\
DeepSeek-Prover-V2-7B & T4 & 8 & +50.0 & [12.5, 87.5] \\
Goedel-Prover-V2-8B & T4 & 8 & +50.0 & [12.5, 87.5] \\
Goedel-Prover-V2-32B & T4 & 7 & +42.9 & [14.3, 85.7] \\
DeepSeek-Prover-V2-7B & T5 & 8 & +12.5 & [-25.0, 50.0] \\
Goedel-Prover-V2-8B & T5 & 8 & +0.0 & [-37.5, 37.5] \\
Goedel-Prover-V2-32B & T5 & 7 & +0.0 & [-42.9, 42.9] \\
\bottomrule
\end{tabular}
\end{table}

\subsection{Mechanism sensitivity}
The mechanism audit contains 185 annotated records and reports unambiguous assignments under the fixed
priority rules used by the analysis. For structural refactorings, the record-level rfl-versus-other
difference is +89.8 points and ranges from +89.1 to +92.6 under leave-one-task-out analysis. At task level,
rfl-bridge proofs survive in 6/6 tasks versus 6/31 for other bridges (Fisher $p=0.0004$); on identical
variants the counts are 9/9 versus 2/9. For semantic refactorings, the record-level rfl-versus-simp difference
is -83.6 points and ranges from -86.2 to -81.0 leave-one-task-out. At task level the counts are 0/3 versus
17/19 (Fisher $p=0.0065$); on identical variants they are 0/4 versus 3/5. Because the rfl groups cover only
six structural and three semantic tasks, these categories are descriptive rather than population-level tactic
effects.

At record level, the three specialized provers contribute 110 structural model--task--variant records, of
which 22 survive. All 12 records with an explicit \texttt{rfl} bridge survive, compared with 10/98 records
without one. The 98 non-bridge records comprise 92 \texttt{simp}/\texttt{dsimp}-style records, four
\texttt{unfold} records, and two other records. Among the 92 \texttt{simp}/\texttt{dsimp}-style records, 10
survive; their residual helper goals close with \texttt{rfl} (2), \texttt{exact} (1), \texttt{aesop} (5), or
\texttt{tauto} (2). Targeted probes confirm that these residual goals are definitionally reducible even when
\texttt{simp [f]} alone does not close them. For semantic refactorings, explicit \texttt{rfl}-equation bridges
survive 0/4 prover records, while \texttt{simp [f]}-style bridges survive 56/67. The T4 contrast is especially
clear: all 24 prover records with a \texttt{simp [f]}-style bridge survive conditional inversion. The T3
failure logs show the complementary brittleness of syntax-specific steps. Among 13 reference T3 failures,
seven are \texttt{rw} pattern mismatches, three stop because \texttt{simp} makes no progress, two leave the
commuted equality unsolved, and one has a term-mode type mismatch. Prover T3 failures likewise include
residual commuted goals and literal \texttt{rfl}, \texttt{exact}, or rewrite steps that expect the original term.

A within-prover prompt intervention also fails to isolate one tactic as a common cause. Forcing an initial
\texttt{unfold f f\_postcond} changes Goedel's structural and semantic PSR by -17.3 and +23.6 points,
respectively, while DeepSeek changes by +5.2 and -13.9 points. The pooled transformation-class interaction is
+4.9 points with 95\% CI [-18.8, 30.6]. We therefore use these tactic categories as descriptive summaries of
whole-proof strategies rather than as causal effects of one tactic.

\subsection{Mechanical repair}
\begin{table}[H]\centering\small
\caption{Predefined one-identifier repair of eligible structural failures. The unedited control reproduces
all 139 attempted failures.}\label{tab:repair}
\begin{tabular}{lrr}
\toprule
Source / template & Repaired & Attempted \\
\midrule
Reference & 51 & 54 \\
DeepSeek-Prover-V2-7B & 21 & 32 \\
Goedel-Prover-V2-8B & 19 & 28 \\
Goedel-Prover-V2-32B & 18 & 25 \\
\midrule
\texttt{dsimp} add helper & 40 & 45 \\
\texttt{simp} add helper & 16 & 37 \\
\texttt{unfold} add helper & 53 & 57 \\
\bottomrule
\end{tabular}
\end{table}
The edit succeeds for 109/139 eligible failures, corresponding to 76.8\% of all 142 structural failures,
with a median change of two tokens on one line. These rates are conditional on template eligibility and do
not measure general proof repair or re-proving.

\subsection{Other sensitivity checks}
Excluding the two variants that require \texttt{noncomputable} changes no headline statistic by more than
0.18 points. For T3, 50/122 candidates certify; 14.0\% of certified candidates are recursive versus 98.6\%
of rejected candidates (Fisher $p=2.2\times10^{-24}$), and median implementation length is 10.5 versus
13.0 lines ($p=0.017$). Re-running the primary
matched percentile bootstrap with 50 random seeds shows discrete endpoint variation for two rows: the
DeepSeek structural lower bound ranges from 0.0 to 2.0, and the Goedel-32B semantic lower bound ranges from
-1.2 to 0.0. We therefore do not treat either class-specific interval as excluding zero. The other seven
matched rows exclude zero under every tested bootstrap seed.

Proof length does not account for the source difference in a task-clustered logistic regression over 269
structural and semantic records. The prover coefficient changes from 1.404 without proof length to 1.370
after adding log proof length ($p=0.0013$), while the proof-length coefficient is 0.030 ($p=0.88$). Among
10--30-line proofs, prover records survive 43.0\% of variants compared with 10.8\% for reference records.
This remains an observational comparison because prompt templates, decoding settings, and FP8 quantization
also differ by source.

\end{document}